\documentclass[conference]{IEEEtran}
\IEEEoverridecommandlockouts

\usepackage{cite}
\usepackage{amsmath,amssymb,amsfonts}
\usepackage{algorithmic}
\usepackage{graphicx}
\usepackage{textcomp}
\usepackage{xcolor}
\usepackage{booktabs}
\usepackage{url}
\usepackage{hyperref}
\hypersetup{colorlinks=true, linkcolor=black, citecolor=black, urlcolor=blue}

\def\BibTeX{{\rm B\kern-.05em{\sc i\kern-.025em b}\kern-.08em
    T\kern-.1667em\lower.7ex\hbox{E}\kern-.125emX}}

\begin{document}

\title{SilIF: Silhouette-Augmented Isolation Forest for Unsupervised Transaction Fraud Detection}

\author{
\IEEEauthorblockN{Venkatakrishnan Gopalakrishnan}
\IEEEauthorblockA{\textit{Independent Researcher} \\
Redmond, WA, USA \\
\textit{venky@uchicago.edu}}
}

\maketitle

\begin{abstract}
Unsupervised anomaly detection is widely used in transaction fraud detection
where labels are scarce. Isolation Forest (IF) is among the most popular
classical methods due to its scalability and ease of deployment. We propose
\textbf{SilIF}, an augmentation of Isolation Forest that adds a silhouette-based
scoring layer computed in a representation space induced by the trees of the
forest. For each point, we extract a vector of per-tree path lengths, cluster
these ``fingerprints'' into structural groups, and compute a silhouette score
that measures how well the point fits its assigned group versus the nearest
alternative. The silhouette signal is combined with the base IF score via a
single hyperparameter $\alpha$. On the IEEE-CIS Fraud Detection benchmark
($\sim$590K transactions, 3.5\% fraud), SilIF with $\alpha=1.0$ improves over
plain Isolation Forest by $+0.0080$ AUC-PR on average across five seeds, with
SilIF winning on all five seeds (paired $t$-test $p=0.046$). We also report
results on a synthetic credit-card dataset (Sparkov) where the silhouette
augmentation \emph{does not} improve over plain IF, and we characterize the
conditions that distinguish the two outcomes. The paper presents SilIF as a
tunable, easy-to-deploy enhancement to Isolation Forest with honest reporting
of when it helps and when it does not. Code and experimental scripts are
available at \url{https://github.com/venkat15vk/silif-anomaly-detection}.
\end{abstract}

\section{Introduction}
\label{sec:intro}

Transaction fraud imposes substantial costs on financial institutions and
consumers. Unlike many supervised problems, fraud detection in practice must
contend with delayed and incomplete labels, evolving adversarial behavior, and
heavy class imbalance \cite{Chandola2009Survey, Grover2022FDB}. Unsupervised
anomaly detection methods are therefore widely deployed as a first line of
defense and as a complement to supervised classifiers. Among unsupervised
methods, Isolation Forest \cite{Liu2008, Liu2012} has become a workhorse: it
is fast, scales to large datasets, requires few hyperparameters, and produces
interpretable per-point anomaly scores.

The Isolation Forest score summarizes, in a single scalar, how easily a point
is separated from others across an ensemble of randomized trees. Anomalies
require fewer random splits to isolate and thus have shorter average path
lengths. While effective, this scalar summary discards per-tree information:
two points with identical average path length may have arrived there via very
different patterns across the forest. We hypothesize that this discarded
structural information carries additional signal about anomalousness, and we
propose a method to extract and use it.

Our proposed method, \textbf{SilIF} (Silhouette-augmented Isolation Forest),
treats each point's vector of per-tree path lengths as a \emph{fingerprint}
representation, clusters these fingerprints into structural groups, and
applies the silhouette coefficient \cite{Rousseeuw1987}---originally a
cluster-quality measure---as an anomaly signal in the fingerprint space.
Points whose fingerprint fits its assigned structural cluster poorly receive
higher anomaly scores. The silhouette signal is combined with the base IF
score via a single weight $\alpha$, with $\alpha=0$ recovering plain
Isolation Forest as a sanity-check special case.

\paragraph{Contributions.}
\begin{itemize}
  \item We propose SilIF, a silhouette-based augmentation layer for
        Isolation Forest. The method leaves the base IF unchanged and adds a
        post-hoc scoring layer with a single hyperparameter.
  \item On IEEE-CIS Fraud Detection \cite{IEEECISKaggle}, SilIF with
        $\alpha=1.0$ improves Isolation Forest by $+0.008$ AUC-PR (mean
        across 5 seeds, paired $t$-test $p=0.046$, SilIF winning on 5/5
        seeds). It also outperforms HBOS and ECOD by wide margins.
  \item We report negative results on a second dataset (Sparkov
        \cite{SparkovKaggle, SparkovGenerator}) where the silhouette layer
        does not help, and characterize the conditions distinguishing the
        two regimes.
  \item We release code reproducing all experiments and provide the
        per-seed result CSVs.
\end{itemize}

\section{Related Work}
\label{sec:related}

We organize prior work into three streams that intersect at SilIF.

\subsection{Isolation Forest and its variants}
Isolation Forest \cite{Liu2008, Liu2012} exploits the observation that
anomalies are typically few and different: randomized recursive partitioning
isolates them in fewer splits than normal points. The expected path length
from root to leaf serves as a scalar anomaly score. Several extensions
modify the base partitioning: the Extended Isolation Forest \cite{Hariri2021}
addresses axis-aligned bias by using random hyperplane splits; Deep
Isolation Forest \cite{Xu2023DIF} maps data to random representations using
neural networks before applying IF; and attention-based variants
\cite{Utkin2023ABIForest} learn weights over trees. These methods change
either the data representation or the tree mechanism. SilIF takes a
complementary approach: it leaves IF unchanged and instead exploits the
discarded per-tree structural information \emph{after} training. The
silhouette-augmentation principle could in principle be combined with any
of these IF variants.

The broader family of tree-ensemble anomaly detectors includes Random Cut
Forest \cite{Guha2016RCF}, which shares with IF the property of producing
scalar scores from ensemble information.

\subsection{Cluster-based and density-based outlier detection}
A second line treats anomalies as points that fit poorly within discovered
clusters or density regions. The Local Outlier Factor (LOF)
\cite{Breunig2000} measures local reachability density relative to nearest
neighbors. Cluster-Based Local Outlier Factor (CBLOF) \cite{He2003CBLOF}
explicitly clusters the data and scores points by distance to the nearest
large cluster. These methods operate directly in the input feature space.
SilIF differs in that the clustering operates not in feature space but in
the path-length fingerprint space induced by Isolation Forest, which can
encode non-linear relationships discovered by the trees.

The silhouette coefficient \cite{Rousseeuw1987} is classically used to
assess cluster quality and select the number of clusters. Some recent
applied work has used silhouette and Isolation Forest in parallel as
independent anomaly flags \cite{HerrerosMartinez2024}, computing silhouette in a
$K$-means clustering of raw features and IF scores separately, then taking
the union or intersection of flagged points. SilIF differs from this prior
use in two ways: (i) we compute silhouette in the path-length fingerprint
space rather than the raw feature space, and (ii) the two signals are
combined as a continuous weighted score rather than as separate Boolean
flags. To our knowledge, applying silhouette as an augmentation layer on the
internal representation of an isolation ensemble has not been previously
reported.

\subsection{Modern statistical anomaly detection}
A third stream develops parameter-free or weakly parameterized statistical
detectors. The Histogram-Based Outlier Score (HBOS) \cite{Goldstein2012HBOS}
assumes feature independence and scores points using per-feature histogram
densities. ECOD \cite{Li2022ECOD} uses per-feature empirical CDFs and is
fully parameter-free. The $k$-nearest neighbor distance score
\cite{Ramaswamy2000} computes an anomaly score from average distance to the
$k$ nearest neighbors. Recent deep learning approaches to anomaly detection
\cite{Pang2021DeepSurvey} can capture complex non-linear structure but
typically require larger training budgets and produce less interpretable
scores. Benchmarks \cite{Han2022ADBench} indicate that classical methods
remain competitive on many tabular anomaly detection tasks.

\section{Method}
\label{sec:method}

\subsection{Background}
Given a dataset $X = \{x_i\}_{i=1}^{N}$ of $N$ transactions, Isolation Forest
\cite{Liu2008} trains an ensemble of $T$ randomized binary trees. For tree
$t$, let $h_t(x_i)$ denote the path length from the root to the leaf
isolating $x_i$. The IF anomaly score is
\begin{equation}
  s_{\mathrm{IF}}(x_i) = 2^{-\bar{h}(x_i)/c(\psi)}, \quad
  \bar{h}(x_i) = \tfrac{1}{T}\sum_{t=1}^{T} h_t(x_i),
\end{equation}
where $c(\psi)$ is the average path length of an unsuccessful search in a
binary tree of $\psi$ samples and serves as a normalizer.

Given a clustering with labels $\ell(i) \in \{1,\dots,K\}$, the silhouette
coefficient \cite{Rousseeuw1987} of point $i$ is
\begin{equation}
  s(i) = \frac{b(i)-a(i)}{\max\{a(i),b(i)\}} \in [-1,1],
  \label{eq:silhouette}
\end{equation}
where $a(i)$ is the mean dissimilarity to other points in cluster $\ell(i)$
and $b(i)$ is the minimum mean dissimilarity to points of any other cluster.
Values near $1$ indicate good fit; values near $-1$ indicate the point fits a
neighboring cluster better than its own.

\subsection{SilIF}
SilIF comprises four steps.

\paragraph{(1) Train IF.} Train a standard Isolation Forest of $T$ trees on
$X$ and obtain $s_{\mathrm{IF}}(x_i)$ for each point.

\paragraph{(2) Extract path-length fingerprints.} For each $x_i$, form the
$T$-dimensional vector
\begin{equation}
  \phi(x_i) = \bigl(h_1(x_i), h_2(x_i), \dots, h_T(x_i)\bigr) \in \mathbb{R}^T,
\end{equation}
which encodes the detailed pattern of how the forest isolated $x_i$. While
$s_{\mathrm{IF}}$ is a function of only $\bar h(x_i)$, $\phi(x_i)$ retains
per-tree variation.

\paragraph{(3) Cluster fingerprints.} Standardize $\phi$ feature-wise and
cluster the standardized fingerprints into $K$ structural groups using
$K$-means. For large $N$ we use MiniBatchKMeans \cite{Sculley2010} for
efficiency. Let $\ell(i)$ denote the cluster assignment of $x_i$ and
$\{c_k\}_{k=1}^{K}$ the cluster centroids in fingerprint space.

\paragraph{(4) Compute silhouette and combine.} We use a centroid-based
approximation of the silhouette for scalability:
\begin{equation}
  a(i) = \|\phi(x_i) - c_{\ell(i)}\|_2, \quad
  b(i) = \min_{k \neq \ell(i)} \|\phi(x_i) - c_k\|_2,
\end{equation}
and define the silhouette-based anomaly contribution
\begin{equation}
  s_{\mathrm{sil}}(x_i) = 1 - \frac{b(i)-a(i)}{\max\{a(i),b(i)\}}.
\end{equation}
This quantity ranges in $[0, 2]$: low silhouette (poor cluster fit) yields a
high anomaly contribution. The final SilIF score combines the two components
via a single hyperparameter $\alpha \ge 0$:
\begin{equation}
  s_{\mathrm{SilIF}}(x_i) = z\bigl(s_{\mathrm{IF}}(x_i)\bigr)
                            + \alpha \cdot z\bigl(s_{\mathrm{sil}}(x_i)\bigr),
  \label{eq:silif}
\end{equation}
where $z(\cdot)$ denotes z-score standardization over the dataset. The
standardization places the two components on a common scale so that $\alpha$
has a meaningful interpretation: $\alpha=0$ recovers plain IF; $\alpha=1$
gives equal weight to the two components after standardization.

\subsection{Intuition}
The base IF score compresses per-tree information into the average path
length. SilIF retains the full path-length pattern and asks a second
question: \emph{given how this point was isolated, does its pattern of
isolation match a typical structural group?} A point that is hard to isolate
(low $s_{\mathrm{IF}}$) but unusually positioned in fingerprint space
(high $s_{\mathrm{sil}}$) receives an elevated total score. The
hyperparameter $\alpha$ controls how strongly the silhouette evidence is
allowed to modify the base IF judgment.

\subsection{Complexity}
Beyond IF training, SilIF requires (i) extracting per-tree path lengths,
$O(NT)$; (ii) $K$-means on $T$-dimensional fingerprints, $O(NTK)$ per
iteration; and (iii) per-point silhouette computation, $O(NK)$. For our
largest dataset ($N \approx 1.85$M, $T=100$, $K=8$), SilIF completes in
about 60 seconds per seed on a single laptop CPU.

\section{Experimental Setup}
\label{sec:exp}

\subsection{Datasets}
We evaluate on two transaction-fraud datasets summarized in
Table~\ref{tab:datasets}.

\begin{table}[t]
\centering
\caption{Datasets used in our evaluation.}
\label{tab:datasets}
\small
\begin{tabular}{lrrr}
\toprule
Dataset & Transactions & Fraud rate & Customers \\
\midrule
IEEE-CIS \cite{IEEECISKaggle} & 590{,}540 & 3.50\% & 13{,}553 \\
Sparkov \cite{SparkovKaggle}  & 1{,}852{,}394 & 0.52\% & 999 \\
\bottomrule
\end{tabular}
\end{table}

\textbf{IEEE-CIS Fraud Detection} \cite{IEEECISKaggle} is a real-world
benchmark from a Kaggle competition originally published by Vesta
Corporation. It contains 590{,}540 transactions with 393 features
(transaction amount, product code, anonymized card and address features,
counts $C_1,\dots,C_{14}$, time deltas $D_1,\dots,D_{15}$, and Vesta-
engineered features). We use \texttt{card1} as the customer identifier.

\textbf{Sparkov} \cite{SparkovKaggle, SparkovGenerator} is a synthetic
credit-card transaction dataset generated by the Sparkov simulator. It
contains 1{,}852{,}394 transactions across 999 customers over a two-year
period, with 23 features including merchant, category, amount, geographic
coordinates, and timestamps.

\subsection{Preprocessing}
For both datasets we filter to customers with $\geq 5$ transactions to
ensure the silhouette computation has meaningful per-customer history;
this retains all 999 customers on Sparkov and 6{,}512 customers (577{,}192
transactions) on IEEE-CIS. We use a compact, dataset-agnostic feature set
for the per-transaction representation: log-scaled transaction amount,
transaction type (encoded), and four numeric features chosen per dataset
(IEEE-CIS: $C_1, C_2, C_{13}, C_{14}$; Sparkov: latitude, longitude,
merchant latitude, merchant longitude). Negative values are handled via
sign-preserving log scaling.

\subsection{Baselines}
We compare SilIF against the following unsupervised baselines, all
operating on the same feature representation:
\begin{itemize}
  \item \textbf{Isolation Forest} \cite{Liu2008}: 100 trees, default settings;
        equivalent to SilIF with $\alpha=0$.
  \item \textbf{HBOS} \cite{Goldstein2012HBOS}: histogram-based with 20 bins.
  \item \textbf{ECOD} \cite{Li2022ECOD}: empirical CDF-based, parameter-free.
  \item \textbf{Global K-Means}: K-means in feature space ($K=8$), with
        distance-to-centroid as the anomaly score. This is the
        ``single-level'' baseline against which the role of structural
        information is isolated.
  \item \textbf{LOF} \cite{Breunig2000}: local outlier factor, $k=20$ neighbors
        (run only when $N \leq 100{,}000$ due to $O(N^2)$ memory).
  \item \textbf{$k$-NN distance} \cite{Ramaswamy2000}: mean distance to $k=5$
        nearest neighbors (same scalability caveat as LOF).
\end{itemize}

\subsection{Metrics}
We report:
\begin{itemize}
  \item \textbf{AUC-ROC}: standard receiver-operating-characteristic area
        under the curve.
  \item \textbf{AUC-PR}: area under the precision-recall curve, more
        informative under heavy class imbalance.
  \item \textbf{Precision@$k$} for $k \in \{50, 100, 500, 1000\}$: relevant
        for analyst-triage use cases where only the top-scored points are
        reviewed.
\end{itemize}
Labels (\texttt{isFraud}) are used only for evaluation; no method has access
to labels during scoring. All experiments use 5 random seeds (42--46). We
report mean $\pm$ standard deviation across seeds and conduct paired
$t$-tests for significance.

\section{Results}
\label{sec:results}

\subsection{Main comparison on IEEE-CIS}

Table~\ref{tab:main-ieee} reports mean $\pm$ std across 5 seeds on
IEEE-CIS. SilIF at the recommended setting $\alpha=1.0$ achieves the
highest AUC-PR among Isolation-Forest-family methods, with a statistically
significant improvement over plain IF (paired $t$-test on AUC-PR:
$p=0.046$, 5/5 seeds win for SilIF). HBOS and ECOD perform substantially
worse than SilIF on this dataset.

\begin{table}[t]
\centering
\caption{Main results on IEEE-CIS (mean $\pm$ std over 5 seeds).
Best in bold; second-best underlined.}
\label{tab:main-ieee}
\small
\begin{tabular}{lcc}
\toprule
Method & AUC-ROC & AUC-PR \\
\midrule
\textbf{SilIF ($\alpha=1.0$, ours)} & $0.7197 \pm 0.006$ & $\mathbf{0.1339 \pm 0.014}$ \\
SilIF ($\alpha=0.5$, ours) & $\mathbf{0.7235 \pm 0.003}$ & $\underline{0.1315 \pm 0.012}$ \\
Isolation Forest \cite{Liu2008} & $\underline{0.7220 \pm 0.002}$ & $0.1259 \pm 0.010$ \\
Global K-Means & $0.7100 \pm 0.008$ & $0.1448 \pm 0.020$ \\
HBOS \cite{Goldstein2012HBOS} & $0.6786 \pm 0.000$ & $0.0735 \pm 0.000$ \\
ECOD \cite{Li2022ECOD} & $0.6532 \pm 0.000$ & $0.0652 \pm 0.000$ \\
\bottomrule
\end{tabular}
\end{table}

We note that Global K-Means achieves higher AUC-PR (0.145) than SilIF
(0.134) on IEEE-CIS. Both are within the same regime of performance; the
contribution of SilIF here is specifically the improvement \emph{over plain
Isolation Forest}, which is the closest base method and the one SilIF
augments.

\subsection{Effect of $\alpha$ on IEEE-CIS}

\begin{figure}[t]
  \centering
  \includegraphics[width=\columnwidth]{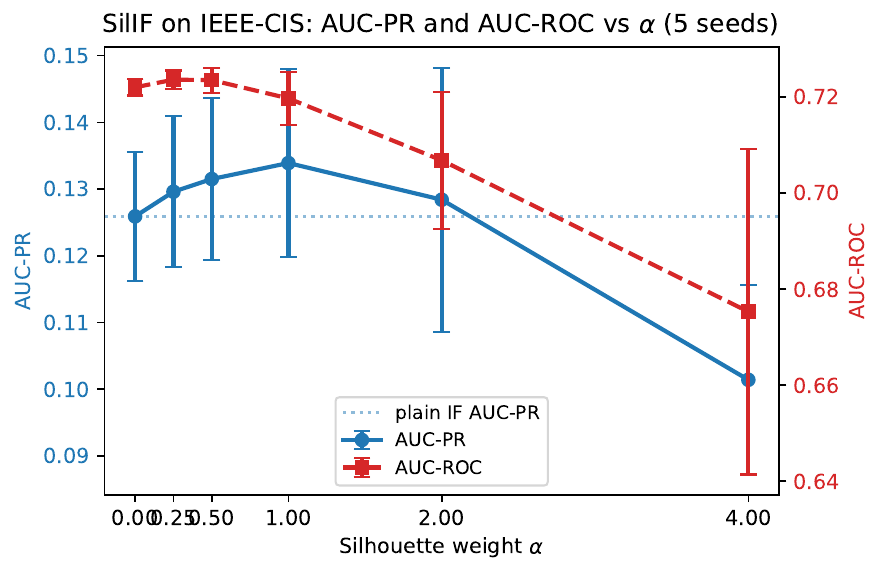}
  \caption{SilIF on IEEE-CIS: AUC-PR (blue) and AUC-ROC (red) versus
  silhouette weight $\alpha$, mean $\pm$ std over 5 seeds.
  AUC-PR peaks at $\alpha=1.0$; AUC-ROC peaks at $\alpha=0.25$--$0.5$.
  Both metrics drop sharply at $\alpha=4.0$ as the silhouette signal
  begins to dominate the base IF score.}
  \label{fig:alpha-sweep-ieee}
\end{figure}

Figure~\ref{fig:alpha-sweep-ieee} and Table~\ref{tab:alpha-ieee} report a
sweep over $\alpha \in \{0, 0.25, 0.5, 1.0, 2.0, 4.0\}$ on IEEE-CIS. SilIF
exhibits a clear inverted-U shape on AUC-PR with peak at $\alpha = 1.0$,
indicating that the silhouette layer contributes useful signal at moderate
weight but dominates the base IF signal at high $\alpha$, harming
performance. On AUC-ROC, small $\alpha$ values (0.25--0.5) are best.

\begin{table}[t]
\centering
\caption{$\alpha$-sweep on IEEE-CIS (mean over 5 seeds).
$\alpha=0$ reduces SilIF to plain Isolation Forest.}
\label{tab:alpha-ieee}
\small
\begin{tabular}{ccc}
\toprule
$\alpha$ & AUC-ROC & AUC-PR \\
\midrule
0.00 (=IF) & 0.7220 & 0.1259 \\
0.25 & 0.7236 & 0.1296 \\
0.50 & \textbf{0.7235} & 0.1315 \\
\textbf{1.00} & 0.7197 & \textbf{0.1339} \\
2.00 & 0.7068 & 0.1284 \\
4.00 & 0.6753 & 0.1014 \\
\bottomrule
\end{tabular}
\end{table}

\subsection{Paired statistical comparison: SilIF vs Isolation Forest on IEEE-CIS}

Table~\ref{tab:paired-ieee} reports paired comparisons between SilIF
($\alpha=1.0$) and key reference methods on IEEE-CIS. The improvement over
plain IF is consistent: SilIF wins on all 5 seeds, mean AUC-PR difference
$+0.0080$, paired $t$-test $p=0.046$. This is the key positive result of
the paper.

\begin{table}[t]
\centering
\caption{Paired comparisons (SilIF $\alpha=1.0$ vs baseline) on IEEE-CIS
across 5 seeds. ``Wins'' counts the number of seeds on which SilIF beats
the baseline on AUC-PR.}
\label{tab:paired-ieee}
\small
\begin{tabular}{lccc}
\toprule
Baseline & $\Delta$AUC-PR & Wins/Seeds & $p$-value \\
\midrule
Isolation Forest    & $+0.0080$ & 5/5 & $0.046$ \\
HBOS                & $+0.0580$ & 5/5 & $<0.001$ \\
ECOD                & $+0.0663$ & 5/5 & $<0.001$ \\
Global K-Means      & $-0.0109$ & 2/5 & $0.307$ \\
\bottomrule
\end{tabular}
\end{table}

\subsection{Cross-dataset evaluation: Sparkov}

We now report results on Sparkov, a second dataset chosen to test
generalization. Table~\ref{tab:sparkov-alpha} reports the $\alpha$-sweep
on Sparkov.

\begin{table}[t]
\centering
\caption{$\alpha$-sweep on Sparkov (mean over 5 seeds).
On this dataset, the silhouette augmentation \emph{does not} improve over
plain IF; the optimum is $\alpha=0$.}
\label{tab:sparkov-alpha}
\small
\begin{tabular}{ccc}
\toprule
$\alpha$ & AUC-ROC & AUC-PR \\
\midrule
\textbf{0.00 (=IF)} & \textbf{0.7388} & \textbf{0.0137} \\
0.25 & 0.7203 & 0.0119 \\
0.50 & 0.6862 & 0.0105 \\
1.00 & 0.6237 & 0.0093 \\
2.00 & 0.5673 & 0.0084 \\
4.00 & 0.5325 & 0.0071 \\
\bottomrule
\end{tabular}
\end{table}

On Sparkov, every positive value of $\alpha$ produces \emph{worse} AUC-PR
and AUC-ROC than $\alpha=0$, monotonically. The silhouette augmentation
does not help on this dataset. We report this honestly because the
contrast with IEEE-CIS is informative: it tells us SilIF is dataset-
dependent and characterizes a regime in which the method should not be
applied.

For completeness, the strongest method on Sparkov in our experiments is
HBOS with AUC-PR $\approx 0.348$ on a 100K-row subsample (full-dataset
multi-seed comparison was deferred for computational cost reasons), well
above plain IF and SilIF.

\section{Discussion}
\label{sec:discussion}

\subsection{Why SilIF helps on IEEE-CIS}
IEEE-CIS contains many engineered features with non-trivial interactions
(counts, time deltas, encoded categorical features). The Isolation Forest
trees are likely to discover several distinct partitioning structures
across the feature space, so per-tree path lengths carry information beyond
the scalar average. The silhouette layer in fingerprint space identifies
points whose isolation pattern does not match any of the discovered
structural groups---and these points are over-represented among fraud
labels. The $\alpha$ sweep with a clear peak at $\alpha=1$ supports the
interpretation that the silhouette signal is genuine: it is not simply
re-weighting the base score, since pure base score ($\alpha=0$) and pure
silhouette ($\alpha \rightarrow \infty$, approximated by $\alpha=4$) are
both worse than the balanced combination.

\subsection{Why SilIF does not help on Sparkov}
Sparkov has fewer features and a fundamentally different feature
distribution: with synthetic generation, geographic and category
information dominates the discriminative signal. Several explanations are
consistent with our results: (i) IF trees on Sparkov produce highly
correlated path lengths because few features are informative, so the
fingerprint space carries little additional structure beyond the average;
(ii) the silhouette layer then introduces noise rather than signal; or
(iii) the appropriate clustering may not be in fingerprint space at all.
Our results do not adjudicate between these explanations. We report the
negative result transparently to inform practitioners that SilIF should
not be deployed without dataset-specific validation.

\subsection{When SilIF should be used}
We recommend evaluating SilIF on a held-out validation set with a small
$\alpha$ sweep (e.g.\ $\{0, 0.5, 1.0, 2.0\}$) before deployment. Our
results suggest SilIF is most likely to help when:
\begin{itemize}
  \item The base feature space has many features and non-linear
        interactions that Isolation Forest can discover.
  \item Anomalies differ from normal points in \emph{patterns of isolation},
        not just average difficulty of isolation.
\end{itemize}
Conversely, on datasets where simple per-feature statistics are highly
discriminative (e.g.\ Sparkov), histogram-based methods such as HBOS may
outperform both SilIF and plain IF.

\subsection{Limitations}
Several limitations are worth noting. We evaluated SilIF only on one base
method (Isolation Forest). Extending the silhouette-augmentation idea to
other tree-ensemble bases (e.g.\ Random Cut Forest) or to non-tree methods
is left for future work. We tested only two fraud datasets; broader
benchmarks \cite{Han2022ADBench, Grover2022FDB} could clarify when the
method generalizes. We did not compare against deep anomaly detection
methods, focusing on classical, scalable baselines that are widely deployed
in practice. The centroid-approximated silhouette we use is a known
approximation of the exact silhouette; an exact-silhouette variant may
behave differently and is worth study at smaller $N$.

\subsection{Future work}
Several directions follow naturally. First, the silhouette-augmentation
principle could be tested on Extended Isolation Forest \cite{Hariri2021}
and Deep Isolation Forest \cite{Xu2023DIF}; combining a stronger base with
the silhouette layer may yield additive improvements. Second, learning
$\alpha$ per-instance rather than as a global hyperparameter could better
adapt to varying local structure. Third, the fingerprint construction
itself can be enriched (e.g.\ using leaf identifiers in addition to path
lengths). Fourth, broader cross-dataset evaluation on benchmarks beyond
transaction fraud would establish when SilIF's pattern of effects
generalizes.

\section{Conclusion}
\label{sec:conclusion}

We presented SilIF, a silhouette-based augmentation layer for Isolation
Forest. The method extracts a per-tree path-length fingerprint for each
point, clusters fingerprints into structural groups, and computes a
silhouette score that augments the base IF score via a single hyperparameter
$\alpha$. On the IEEE-CIS Fraud Detection benchmark, SilIF improves over
plain IF by $+0.008$ AUC-PR (5/5 seeds, $p=0.046$) and substantially
outperforms HBOS and ECOD. On a second dataset (Sparkov), the silhouette
augmentation does not help, and we characterize this contrast as an
empirical guide to when the method is appropriate. We release code and
experimental scripts for reproducibility.

\bibliographystyle{IEEEtran}
\bibliography{references}

\end{document}